\documentclass[10pt,twocolumn,letterpaper]{article}

\usepackage{wacv}
\usepackage{times}
\usepackage{epsfig}
\usepackage{graphicx}
\usepackage{amsmath}
\usepackage{amssymb}
\usepackage{booktabs}

\usepackage{enumitem}
\usepackage{verbatim}
\usepackage{pifont}
\usepackage{dblfloatfix}
\usepackage{footmisc}
\usepackage{graphicx}
\usepackage{booktabs}
\usepackage{float}
\usepackage{dblfloatfix}
\newcommand{\cmark}{\ding{51}}%
\newcommand{\xmark}{\ding{53}}%

\def\wacvPaperID{43} 

\wacvfinalcopy 

\ifwacvfinal
\usepackage[breaklinks=true,bookmarks=false]{hyperref}
\else
\usepackage[pagebackref=true,breaklinks=true,colorlinks,bookmarks=false]{hyperref}
\fi

\begin{document}

\title{ Towards Online Domain Adaptive Object Detection}

\author{Vibashan VS, Poojan Oza, and Vishal M. Patel \\
 Johns Hopkins University, Baltimore, MD, USA \\
{\tt\small \{vvishnu2,poza2,vpatel36\}@jhu.edu}
}

\maketitle
\thispagestyle{empty}

\begin{abstract}
   Existing object detection models assume both the training and test data are sampled from the same source domain. This assumption does not hold true when these detectors are deployed in real-world applications, where they encounter new visual domains. Unsupervised Domain Adaptation (UDA) methods are generally employed to mitigate the adverse effects caused by domain shift. Existing UDA methods operate in an offline manner where the model is first adapted toward the target domain and then deployed in real-world applications. However, this offline adaptation strategy is not suitable for real-world applications as the model frequently encounters new domain shifts. Hence, it is critical to develop a feasible UDA method that generalizes to the new domain shifts encountered during deployment time in a continuous online manner.
   To this end, we propose a novel unified adaptation framework that adapts and improves generalization on the target domain in both offline and online settings. Specifically, we introduce MemXformer - a cross-attention transformer-based memory module where items in the memory take advantage of domain shifts and record prototypical patterns of the target distribution. Further, MemXformer produces strong positive and negative pairs to guide a novel contrastive loss, which enhances target-specific representation learning.
   Experiments on diverse detection benchmarks show that the proposed strategy producs state-of-the-art performance in both offline and online settings.
   To the best of our knowledge, this is the first work to address online and offline adaptation settings for object detection. Source code:  \href{https://github.com/Vibashan/memXformer-online-da}{https://github.com/Vibashan/memXformer-online-da}
\end{abstract}
\vspace{- 2.5 em}
\section{Introduction}
\label{sec:intro}

The ability to train deep network models on large-scale annotated datasets \cite{krizhevsky2012imagenet,everingham2010pascal,lin2014microsoft,johnson2016driving} has accelerated the progress for multiple computer vision tasks such as classification \cite{krizhevsky2012imagenet,he2016deep,dosovitskiy2020image}, segmentation \cite{long2015fully,zhao2017pyramid,sun2019high}, and detection \cite{ren2015faster,redmon2016you,liu2016ssd}.
Despite this success, these models have limited generalization capabilities \cite{szegedy2013intriguing,hendrycks2019benchmarking,geirhos2018generalisation}.
Specifically, the model performance drops when the test data (target domain) is sampled from a different distribution than that of the training data (source domain) \cite{ben2010theory}.
For example, when a model is deployed in real-world applications such as autonomous navigation, it could encounter images with weather-based degradations, camera artifacts, etc., unknown during training.

\begin{figure}[t!]
    \begin{center}
        \includegraphics[width=1.0\linewidth]{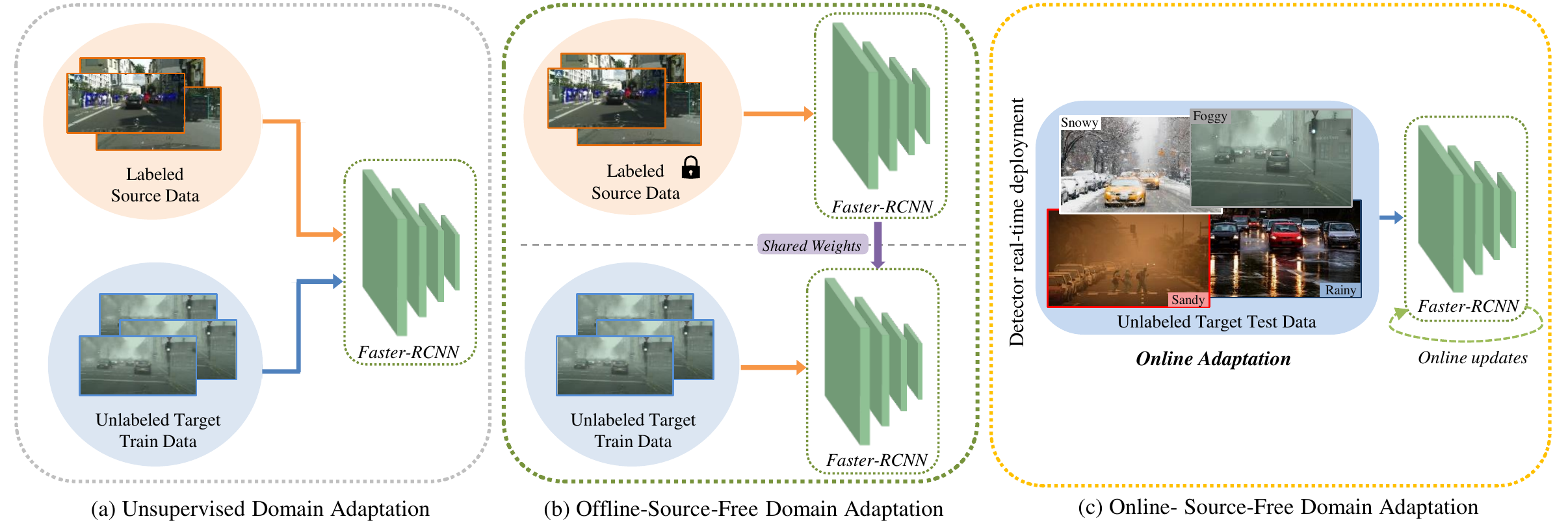}
    \end{center}
    \vskip -10.0pt 
    \caption{\textbf{Left:} Unsupervised Domain Adaptation - labeled source data and unlabeled target data are available during adaptation. \textbf{Middle:} Source-Free Domain Adaptation - source-trained model is adapted to the target domain. \textbf{Right:} Online Source-Free Domain Adaptation - source-trained model is adapted to target distribution shift during real-world deployment via online updates. }
    \label{fig:intro} 
    \vskip -17.0pt
\end{figure}

Unsupervised Domain Adaptation (UDA) methods \cite{ganin2016domain,tzeng2017adversarial,saito2018maximum,chen2017no,hoffman2016fcns,hoffman2018cycada,chen2018domain,inoue2018cross,saito2019strong} are generally employed to improve model generalization under domain shift condition.
Existing UDA methods assume that both labelled source data and unlabeled target data are available during adaptation. This scenario is often not feasible in current real-world applications, as the labelled source data is often restricted due to privacy regulations, data transmission constraints, or proprietary data concerns.
To overcome this drawback, recently, some works have explored Source-Free Domain Adaptation (SFDA) \cite{kundu2020universal,kim2021domain,xia2021adaptive,liang2020we,li2020free} setting, where a source-trained model is adapted towards the target domain without requiring access to the source data. 
However, in both UDA and SFDA settings, adaptation is performed in an offline manner where the model is first adapted towards the target domain and then deployed in real-world applications.
In addition, it is often impossible to have prior knowledge about the target domain in most real-world applications.
In other words, the deployed model could encounter a diverse set of target domains and offline adaptation to every distribution shift would be infeasible. Therefore, we propose a unified adaptation framework which utilizes a source-trained detector and adapts to the target domain in both offline and online manner.

In recent years, few works have explored various test-time adaptation settings where adaptation is performed during test-time \cite{wang2021tent, yao2021cross, valanarasu2022fly}. Wang \cite{wang2021tent} proposed a fully test-time adaptation strategy which performs entropy minimization during test-time and only updates the model batch-norm parameters for the classification task. However, extending TENT  to detection framework \cite{wang2021tent} has two critical drawbacks: 1) TENT use a very large batch size during test-time adaptation, which is not feasible during real-time deployment as images arrive one by one sequentially. 2) Updating only the batch norm parameter of a network with batch size 1 essentially degrades the model performance \cite{yao2021cross}. Although existing test-time adaptation settings are closer to online-SFDA settings, they are not suitable for adapting a detection model during real-world deployment. To overcome these issues, we explore an Online Source-Free Domain Adaptation (Online-SFDA)setting, where a model is adapted to any distribution shifts encountered during deployment in an online manner with batch size 1. Fig.~\ref{fig:intro} illustrate the online source-free domain adaptation setting for detection and its differences against the other adaptation settings.

Source-free domain adaptive object detection is relatively new and a more challenging setting than UDA. Existing SFDA methods \cite{li2020free,huang2021model} for detection adapt to the target domain by training on the pseudo-labels generated by the source-trained model. Due to domain shifts, these generated pseudo-labels are noisy and training a model on top of them would lead to noise overfitting \cite{liu2021unbiased,deng2021unbiased}. To alleviate these issues, we employ a mean-teacher framework where the student model is supervised using pseudo-labels generated by the teacher network and the teacher network is slowly updated via the exponential moving average (EMA) of student weights. Therefore, the student network is trained on consistent pseudo-labels leading to less overfitting and the teacher network is a gradual ensemble of target adapted student weights \cite{liu2021unbiased}. However, this strategy is inefficient in learning two critical aspects required for optimal online adaptation: 1) They fail to learn robust target feature representation, 2) They fail to fully exploit the online target samples. Hence, we propose a novel memory module and a contrastive loss to fully utilize online target samples and learn robust target feature representation.

Contrastive Learning (CL) \cite{chen2020simple,chen2020big,he2020momentum,chopra2005learning,khosla2020supervised} aims to learn high-quality features from unlabeled data by forcing similar object instances to stay close and push dissimilar ones apart in an unsupervised manner.
This is especially useful for online-SFDA as source-labelled data are unavailable during adaptation. 
Existing CL methods are designed for classification tasks where they operate on image-level features and require multiple image views (or augmentations) \cite{chen2020simple} to learn robust feature representation.
Consequently, obtaining these large sets of views through input augmentations is computationally expensive for adapting detector models.
However in detector models, it is possible to obtain different views for an object in an input image without heavy input augmentations.
More precisely, the detector provides multiple object proposals generated by Region Proposal Network (RPN), which in turn provides multiple cropped views around the object instance at different locations and various scales. Therefore, applying CL loss on RPN cropped views guides the model to learn object-level feature representation on the target domain.
Note this CL loss is used to supervise the student network, where the object-level features are obtained from the student RoI features. 
However to perform contrastive learning, these student RoI features require positive and negative pairs. 
To this end, we propose \textit{MemXfromer}, a cross-attention transformer-based memory module where items in the memory record prototypical patterns of the continuous target distribution. The proposed MemXfromer solves two important problems for online adaptation: 1) store the target distribution during online adaptation, which are utilized for future adaptation. 2) stored temporal ensemble of target representations provides positive and negative pairs to guide the contrastive learning process. 
Further, we introduce a cross-attention based read and write technique which models better target distribution and provides strong positive and negative pairs for contrastive learning. Note that the proposed method is not only suitable for online adaptation but also for offline adaptation. 
In a nutshell, this paper makes the following contributions:
\begin{itemize}[topsep=0pt,noitemsep,leftmargin=*]
    \item To the best of our knowledge, this is the first work to consider both online and offline adaptation settings for detector models. 
    \item We propose a novel unified adaptation framework which makes the detector models robust against online target distribution shifts.
    \item We introduce the MemXformer module, which stores prototypical patterns of the target distribution and provides contrastive pairs to boost contrastive learning on the target domain. 
    \item We consider multiple detection benchmarks for experimental analysis and show that the proposed method outperforms existing UDA, and SFDA methods for both online and offline settings.
    \vskip -8 pt
\end{itemize}



\section{Related works}\label{sec:related_works}
\vskip -8 pt
\noindent \textbf{Unsupervised domain adaptation.} Existing unsupervised domain adaptation methods can be categorized into three groups based on adversarial training \cite{Chen2018DomainAF,saito2019strong,Sindagi_DA_Detection_ECCV2020,vs2022meta},  self-training \cite{khodabandeh2019robust,wu2021instance} and image-to-image translation \cite{kim2019diversify,roychowdhury2019automatic}. The first domain adaptive object detection was studied in \cite{Chen2018DomainAF}, where they followed an adversarial-based strategy to perform feature alignment at both image-level and instance-level to mitigate the domain shift. Later, Saito \cite{saito2019strong}  proposed an adversarial-based strategy where strong alignment of the local features and weak alignment of the global features. Kim \cite{kim2019diversify}, introduced an image-to-image translation-based where multiple target domain images are created by stylizing the labelled source images. Multiple discriminators are used to performing adversarial alignment to reduce domain discrepancy by utilizing these target-styled source images. In \cite{khodabandeh2019robust}, a pseudo-label based training strategy was formulated to counter noise in pseudo-labels to perform robust training of object detectors on the target domain. However, all these works assume to have access to labelled source data and unlabeled target data during adaptation, and they operate in an offline setting.
 
\noindent \textbf{Source-free domain adaptation.} In the source-free domain adaptation setting, we have a source-trained model which adapts to the target domain without having access to source data. Multiple works have addressed the source-free domain adaptation (SFDA) setting for classification \cite{liang2020we,li2020model}, segmentation \cite{liu2021source,kundu2021generalize,vs2022target} and object detection \cite{li2020free,huang2021model,vs2022instance,hegde2021attentive,hegde2021uncertainty} tasks. In detail, for classification task \cite{li2020free} proposed a self-supervised method to learn target domain representation via information maximization. Further for segmentation \cite{liu2021source,kundu2021generalize}  and object detection \cite{li2020free,huang2021model}, the proposed methods are based on pseudo-label self-training to learn target-specific representation. However, similar to existing UDA works, these SFDA methods operate in an offline setting. Thus, we explore online adaptation, which is a more practical way to tackle domain shifts for real-world applications. 

\noindent \textbf{Online adaptation.} 
Sun \cite{sun2020test} proposed a Test-time training (TTT) strategy, where a model is trained on source data along with an auxiliary task (eg: rotation prediction) which is utilized during test-time to fine-tune the model on target test distribution. The major drawback of this adaptation strategy is training an auxiliary task along with source training just to perform adaptation during test-time is not a feasible solution and effective solution for real-world application. Later, Wang \cite{wang2021tent} proposed a fully test-time adaptation setting, where the given source trained model adapts to the target domain by entropy minimization during test-time in an online manner by entropy minimization. In this way, Tent \cite{wang2021tent} adapts to the target domain with test-time loss. Here, the major limitation of \cite{wang2021tent} is a requirement of a large batch size during test-time adaptation, which is not feasible during real-time deployment as images arrive one by one sequentially. Although existing test-time adaptation settings provide close resemblance to online-SFDA settings, these test-time settings are not suitable for adapting a detection model during real-world deployment. Therefore in this work, we explore both online and offline adaptation settings for the object detection tasks. 

\noindent \textbf{Contrastive representation learning.} Contrastive representation learning has shown huge progress towards unsupervised feature learning. The standard way of formulating contrastive learning for an anchor is by pulling together the feature embedding of anchor's positive pairs and pushing apart from the anchor's negative pair \cite{oord2018representation,chen2020simple,he2020momentum}. These positive and negative pairs are formed by augmenting the anchor image and sampling from the input batch of images. Thus, for a given anchor, the positive pair are augmented anchor images and the negative pairs are other images from the batch \cite{oord2018representation,chen2020simple,he2020momentum}. On top of this, by exploiting the task-specific label information, \cite{khosla2020supervised} performed contrastive learning in a supervised manner. Nonetheless, all these tasks require a large batch size to perform contrastive learning effectively and it is not feasible to have more than one image during online adaptation. Thus, we propose a memory-based contrastive learning framework suitable for adapting object detectors during deployment in an online manner.


\begin{figure*}[t!]
    \begin{center}
     	\includegraphics[width=0.80\linewidth]{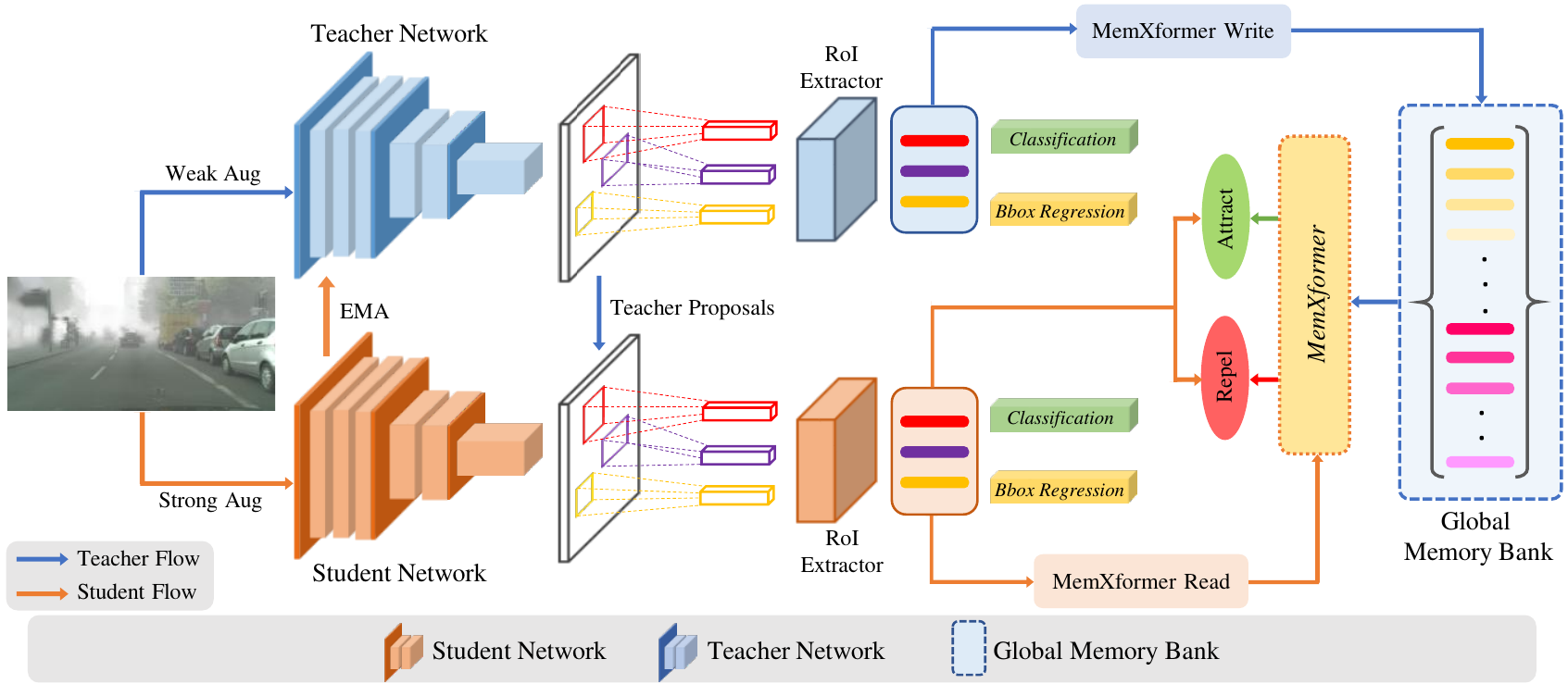}
    \end{center}
     \vskip -8 pt\caption{Overview of the proposed online-SFDA training pipeline. The detection network is adapted to online target distribution shifts by improving target representations through contrastive training. Specifically, the proposed MemXformer records prototypical patterns of the target distribution shift and provides strong positive and negative pairs to guide the contrastive learning process.
distribution .}
\label{fig:proposed_method_blockdiagram}
\vskip -17 pt
\end{figure*}

\vspace{- 1 em}
\section{Proposed method}\label{sec:proposed_method}
\vskip -10 pt
The online-SFDA setting considers a source-trained model with parameters $\Theta_{src}$ and adapts to any target distribution shifts during real-world deployment as illustrated in Fig.~\ref{fig:intro}.
Let us consider a stream of online target data denoted as $\mathcal{T} = \{x_1, x_2,.., x_n\}$, where $x_n$ is the $n^{th}$ online sample.
Since these samples arrive sequentially, the model gets adapted to each sample and the adapted weights are used for future online samples.
Specifically, the model parameters during adaptation on the $n^{th}$ sample $x_n$, i.e. $\Theta_{src}^{(n)}$, are initialized with the model parameters updated through online adaptation of previous $x_{n-1}^{th}$ sample.
To summarize, online-SFDA performs continuous online adaptation, i.e., adaptation will be continued as long as there is a stream of data and necessity.

\noindent \textbf{Student-teacher training.} In online-SFDA, the model parameters need to be continuously updated in an online unsupervised manner.
Consequently, the model risks forgetting the original hypothesis learned through supervised source training \cite{liu2021unbiased,deng2021unbiased}.
To overcome this, prior works \cite{tarvainen2017mean,liu2021unbiased} have employed a student-teacher framework.
Specifically, the student parameters ($\Theta_{std}$) are adapted to the target domain by minimizing the detection loss supervised through the teacher-generated pseudo-labels. The adapted student parameters are then transferred to the teacher parameters ($\Theta_{tch}$) via Exponential Moving Average (EMA).
This can be formally written as:
\setlength{\belowdisplayskip}{1pt} \setlength{\belowdisplayshortskip}{1pt}
\setlength{\abovedisplayskip}{1pt} \setlength{\abovedisplayshortskip}{1pt}
\begin{align}
\label{eq:teach_up}
& \mathcal{L}_{pl}(x_n) = \mathcal{L}_{rpn}(x_{n},\tilde{y}_{n})+\mathcal{L}_{rcnn}(x_{n},\tilde{y}_{n})\\
& \Theta^{(n+1)}_{std} \leftarrow \Theta^{(n)}_{std} + \gamma \frac{\partial(\mathcal{L}_{pl}(x_n))}{\partial \Theta^{(n)}_{std}} \\
& \Theta^{(n+1)}_{tch} \leftarrow \alpha \Theta^{(n)}_{tch}+(1-\alpha)\Theta^{(n+1)}_{std},
\end{align}
where $x_n$ and $\tilde{y}_{n}$ are the $n^{th}$ test sample and corresponding pseudo-label generated by teacher network, $\mathcal{L}_{pl}$ is the pseudo-label supervision loss, $\gamma$ is the student learning rate, and $\alpha$ is teacher EMA rate.
However, the student-teacher framework is still not sufficient to learn robust features to mitigate target distribution shifts.
Hence, we explore contrastive learning-based strategies further to improve the robustness of feature representations in an online setting.

\noindent \textbf{Contrastive Learning (CL).} SimCLR \cite{chen2020simple} is a commonly used CL framework which learns representations for an image by maximizing agreement between differently augmented views of the same sample.
For given an anchor image $x_i$, the SimCLR loss can be written as:
\begin{equation}
\label{eq:simclr}
\mathcal{L}_{\text{SimCLR}}(x_i) = - {\log}\left( \frac{\operatorname{exp}(\operatorname{sim}(z_i, z_j))}{\sum_{l=1, \ni_{l\neq i}}^{2N}  \operatorname{exp}(\operatorname{sim}(z_i, z_l))} \right),
\end{equation}
where $N$ is the batch size, $z_i$ and $z_j$ are the features of two different augmentations of the same sample $x_i$, whereas $z_l$ represents the feature of the $l^{th}$ batch sample $x_l$, where $l \neq i$.
Also, $\operatorname{sim}(\cdot, \cdot)$ indicates a similarity function, e.g. cosine similarity. 
Note that in general, the CRL framework assumes that each image contains one category/object \cite{chen2020simple}.
Moreover, it requires large batch sizes that could provide multiple positive/negative pairs for training \cite{chen2020big}.
In contrast for object detection, each image will have multiple objects and a large batch size or multiple views are computationally not feasible.
Hence, existing CRL methods are more suited for classification tasks.

\subsection{Memory-based contrastive learning}
Though existing contrastive learning methods like SimCLR are exceptional at learning high-quality representations, they are more suitable for the classification task. For detection, these CL methods require large batch size and heavy input augmentation, which are computationally expensive to apply for online parameter updates (discussed in Sec.~\ref{sec:intro}).
Therefore, we utilize a computationally efficient memory-based approach to make contrastive learning feasible and effective for online model updates.
The proposed online-SFDA strategy is illustrated in Fig.~\ref{fig:proposed_method_blockdiagram}.

\begin{figure}[t!]
	\begin{center}
		\includegraphics[width=1.0\linewidth]{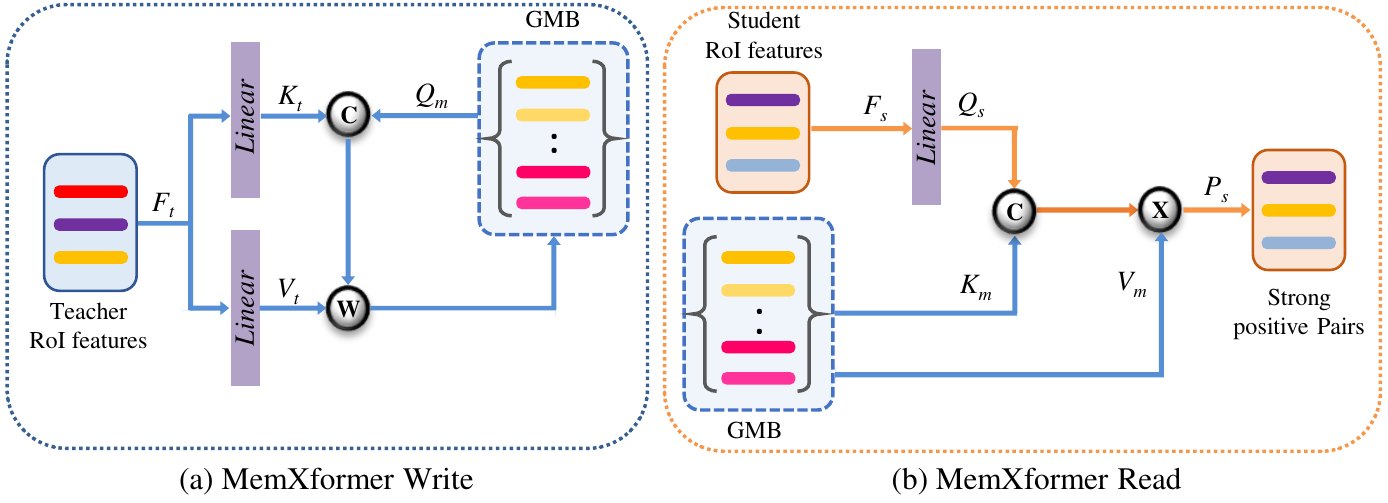}
	\end{center}
	\vskip-10.0pt\caption{MemXformer Write and Read operations.}
	\label{fig:mem_trans} 
	\vskip-20.0pt
\end{figure}

\noindent \textbf{MemXformer.} A cross-attention transformer-based memory module which stores target distribution shift and guides the contrastive learning for target domain representation during online adaptation. Specifically, we employ a Global Memory Bank $M = \{m^i \in \mathbb{R}^{1 \times C}\}_{i=1}^{N_{l}}$,  where $N_l$ is number of memory items and $C$ is memory item feature dimension. These memory items are used to store target representation and record prototypical patterns of the target distribution during the adaptation. In addition, these memory items are used to retrieve strong positive and negative pairs for guiding contrastive learning. The MemXformer module has two operations: write and read, which are based on cross-attention. In the MemXformer write operation, the teacher RoI features are used to update the memory elements appropriately. In the MemXformer read operation, the student RoI features are queried to the memory and a weighted sum of similar memory elements is retrieved, which essentially provide strong positive pairs. The read and write operations of MemXformer are illustrated in Fig.~\ref{fig:mem_trans}.

\noindent \textbf{Write.} To update the memory elements, we consider only the teacher network RoI features $\mathcal{F}_{t} = \{f^{i}_{t} \in \mathbb{R}^{1 \times C}\}_{i=1}^{N_{f}}$, where $N_{f}$ is number of RoI features and $C$ is RoI feature dimension. The teacher RoI features are considered because in the student-teacher framework, the teacher pipeline has input with weak augmentations resulting in accurate RPN proposals compared to the student pipeline. As shown in Fig. \ref{fig:mem_trans} (a), first the teacher RoI features are projected as \textit{key} $K_t = \{k^{i}_{t}\}_{i=1}^{N_{f}}$ and \textit{value} $V_t = \{v^{i}_{t}\}_{i=1}^{N_{f}}$ using two FC layer with weight $W_k$ and $W_v$, respectively. Now each memory items are considered as \textit{query} $Q_m = \{m^{j}\}_{j=1}^{N_{l}}$ and we compute a cross-attention map $S_t$ between the teacher RoI features and memory items as follows:
\begin{align}
\label{eqn:eqlabel}
\begin{split}
{k}_{t}^{i} &= W_k \cdot {f}_{t}^{i},
\\
{v}_{t}^{i} &= W_v \cdot {f}_{t}^{i},
\\
\end{split}
\end{align}
\begin{align}\label{eq:update_memory_1}
s^{(i,j)}_{t} &= \frac{\exp \left({m}^{j}\left({k}_{t}^{i}\right)^T\right)}{\sum_{l \in M} \exp \left( {m}^{l}\left(  {k}_{t}^{i}\right)^T\right)},
\end{align}
where the cross-attention map $S_t$ is a 2D matrix of size $N_m \times N_f$ and $s_{t}^{i,j}$ represents how $j^{th}$ memory items is related to $i^{th}$ teacher RoI features. We utilize this cross-attention map $S_t$ and $V_t$ to update $j^{th}$ memory item using following equation: 
\begin{equation}\label{eq:update_memory_2}
{m}^{j} \leftarrow F \left({m}^{j}+\sum_{i \in V} s^{(i, j)}_{t} {v}_{t}^{i}\right).
\end{equation}
 where $F(.)$ is $L_2$ norm. Therefore, using attention-based weighted average and global memory bank update for each online sample makes the MemXformer effectively store and model the target distribution. 

\noindent \textbf{Read.} To read the memory elements, we consider only the student network RoI features $\mathcal{F}_{s} = \{f^{i}_{s} \in \mathbb{R}^{1 \times C}\}_{i=1}^{N_{f}}$, where $N_f$ is number of RoI features and $C$ is RoI feature dimension. In addition, the MemXformer Read operation is performed to obtain strong positive pairs given student RoI features as a query. As shown in Fig. \ref{fig:mem_trans} (b), first the student RoI features are projected as \textit{query} $Q_s = \{q^{i}_{s}\}_{i=1}^{N_{f}}$ by one FC layer with weight $W_q$. Now each memory items are considered as \textit{key} $K_m = \{{m}^{j}\}_{j=1}^{N_{l}}$ and we compute a cross-attention map $S_s$ between the student RoI features and memory items as follows:
\begin{align}
\label{eqn:eqlabel}
\begin{split}
{q}_{s}^{i} &= W_k \cdot {f}_{s}^{i},
\end{split}
\end{align}
\begin{align}\label{eq:update_memory_1}
s^{(i,j)}_{s} &= \frac{\exp \left({q}_{s}^{i} \left({m}^{j}\right)^T\right)}{\sum_{l \in M} \exp \left({q}_{s}^{i} \left( {m}^{l}\right)^T\right)},
\end{align}
where the cross-attention map $S_s$ is a 2D matrix of size $N_m \times N_f$ and given $i^{th}$ student RoI features as query, the  $s_{t}^{i}$ $th$ row presents $N_l$ memory items attention score. Therefore, given $i^{th}$ student RoI features as query, we generate its corresponding positive pair by attention guided weighted sum of most similar memory items. Thus, utilizing the cross-attention map $S_s$ and considering memory items as \textit{value} $V_m = \{m^{j}\}_{j=1}^{N_{l}}$, we compute the strong positive pair for $i^{th}$ student RoI features using following equation: 
\begin{equation}\label{eq:update_memory_2}
{p}^{i}_{s} = \sum_{j \in M} s^{(i, j)}_{s} {m}^{j}.
\end{equation}
where $\mathcal{P}_{s} = \{p^{i}_{s}\}_{i=1}^{N_{f}}$ corresponds to set of strong positive pair for student RoI features $\mathcal{F}_s$. In detail, the retrieved positive pairs are temporal ensembles of the prototypical target distribution, which gives more information regarding the online target distribution shifts. This essentially guides contrastive learning to model the target distribution. 

\noindent \textbf{Negative Pair Mining.} As explained earlier from MemXformer read operation, we obtain a set of strong positive pairs for a given student RoI feature. These strong positive pairs are essentially an ensemble of most similar memory items. However, these ensembled similar memory items also contain dissimilar memory items but are scaled with less attention weights. This restricts the contrastive learning capability to effectively model the target domain representation. To mitigate the dissimilar item's effect on CL, we propose negative pair mining. Specifically in negative pair mining, given a student RoI feature as query and cross-attention map $S_s$, we mine the least similar 10$\%$ of the memory items and label them as negative pairs $\mathcal{N}_s = \{m_{i}^{n}\}_{i=1}^{N_{s}}$. As a result, by performing negative pair mining, we obtain $N_s$ negative samples for one positive sample, where $N_s$ is top 10$\%$ of least similar memory items.

\noindent \textbf{Memory contrastive loss.} Given student RoI feature $f_{s}^{i}$ as anchor, utilizing MemXformer Read operation and negative pair mining we obtain strong positive $\mathcal{P}_{s}$ and negative pairs $\mathcal{N}_{s}$ from MemXformer. Therefore, given an image $x_n$ with student RoI feature $\mathcal{F}_{s}$, the MemCLR loss is calculated as: 
\begin{small}
\label{eq:mem_clr}
\begin{math}
 \mathcal{L}_{\text{MemCLR}}({x}_{n}) = \\ -\log\left\{   \frac{1}{\left| \mathcal{F}_{s} \right|} \sum_{i\in \mathcal{F}_{s}} \frac{\operatorname{exp}(f_{s}^{i}  \cdot p_{s}^{i})}{ \operatorname{exp}(f_{s}^{i}  \cdot p_{s}^{i}) +   \sum_{n\in \mathcal{N}_{s}} \operatorname{exp}(f_{s}^{i} \cdot {m}^{n} )}\right\},
\end{math}
\end{small}\\
Therefore, minimizing the MemCLR loss guided by strong positive and negative pairs enhance the student model to learn better target representation in an online-SFDA setting.

\noindent \textbf{Overall loss.} We illustrate our overall architecture for online source-free domain adaptation in Fig.~\ref{fig:proposed_method_blockdiagram}.
The proposed method utilizes a global memory bank to perform memory-based contrastive learning to robustify the representations under varying target distribution shifts.
Therefore, the overall online-SFDA loss for any online sample $x_n$ can be calculated as:
\begin{equation}
\mathcal{L}_{FTTA}(x_n) \ = \ \mathcal{L}_{pl}^{st}(x_n) \ + \  \mathcal{L}_{\text{MemCLR}}(x_n). \nonumber
\end{equation}
\vspace{- 2 em}


\begin{table*}[!t]
\caption{Quantitative results (mAP) for Cityscapes $\rightarrow$  FoggyCityscapes. S: Source-only, O: Oracle, UDA: Unsupervised Domain Adaptation, SFDA: Source-Free Domain Adaptation, O-SFDA: Online Source-Free Domain Adaptation.}
\label{tab:foggy}
\centering
\resizebox{0.85\linewidth}{!}{\begin{tabular}{clccccccccccc}
\hline
Type  & Method    & Offline & Online     & prsn          & rider         & car           & truck         & bus           & train         & mcycle        & bicycle       & mAP           \\ \hline
S  & Source-Only  & \cmark & \xmark   & 29.3          & 34.1          & 35.8          & 15.4          & 26.0          & 9.09          & 22.4          & 29.7          & 25.2          \\ \hline
 & DA Faster \cite{Chen2018DomainAF} (CVPR 2018) & \cmark & \xmark  & 25.0          & 31.0          & 40.5          & 22.1 & 35.3          & 20.2          & 20.0          & 27.1          & 27.6          \\ 
 & Selective DA \cite{zhu2019adapting} (CVPR 2019) & \cmark & \xmark &  33.5& 38.0& 48.5& 26.5& 39.0& 23.3& 28.0& 33.6& 33.8          \\
 & D\&Match \cite{kim2019diversify} (CVPR 2019) & \cmark & \xmark  & 30.8& 40.5& 44.3& 27.2& 38.4& 34.5& 28.4& 32.2& 34.6\\
 UDA & MAF \cite{he2019multi} (ICCV 2019) & \cmark & \xmark  & 28.2& 39.5& 43.9& 23.8& 39.9& 33.3& 29.2& 33.9& 34.0\\
 & Robust DA \cite{khodabandeh2019robust} (ICCV 2019) & \cmark & \xmark  & 35.1& 42.1& 49.1& 30.0& 45.2& 26.9& 26.8& 36.0& 36.4\\
 & MTOR \cite{cai2019exploring} (CVPR 2019) & \cmark & \xmark  & 30.6& 41.4& 44.0& 21.9& 38.6& 40.6& 28.3& 35.6& 35.1\\
 & Strong-Weak \cite{saito2019strong} (CVPR 2019)& \cmark & \xmark  & 29.9& 42.3& 43.5& 24.5& 36.2& 32.6& 30.0& 35.3& 34.3\\ 
 & Categorical DA \cite{xu2020exploring} (CVPR 2020) & \cmark & \xmark & 32.9& 43.8& 49.2& 27.2& 45.1& 36.4& 30.3& 34.6& 37.4\\
    & MeGA CDA \cite{hsu2020progressive} (CVPR 2021) & \cmark & \xmark& 37.7& 49.0& 52.4& 25.4& 49.2& 46.9& 34.5& 39.0& 41.8\\
    & Unbiased DA \cite{deng2021unbiased} (CVPR 2021) & \cmark & \xmark & 33.8& 47.3& 49.8& 30.0& 48.2& 42.1& 33.0& 37.3& 40.4\\\hline
  & SFOD \cite{li2020free} (AAAI 2021) & \cmark & \xmark  & 25.5& 44.5& 40.7& \textbf{33.2}& 22.2& 28.4& 34.1& 39.0& 33.5\\ 
SFDA & HCL \cite{huang2021model} (NeurIPS 2021) & \cmark & \xmark  & 26.9& \textbf{46.0}& 41.3& 33.0& 25.0& 28.1& \textbf{35.9}& 40.7& 34.6\\ 
& Mean-Teacher \cite{tarvainen2017mean} & \cmark & \xmark  & 33.9  & 43.0& 45.0& 29.2& 37.2& 25.1& 25.6& 38.2 & 34.3  \\
& MemCLR (Ours)  & \cmark & \xmark   & \textbf{37.7}& 42.8& \textbf{52.4}& 24.5& \textbf{40.6}& \textbf{31.7}& 29.4& \textbf{42.2}& \textbf{37.7}  \\ \hline
  O-SFDA & Tent \cite{wang2020tent} (ICLR 2021) & \xmark & \cmark    & 31.2& 38.6& 37.1& 20.2& 23.4& 10.1& 21.7& 33.4& 26.8 \\ 
 & MemCLR (Ours)  & \xmark & \cmark   & \textbf{32.1}& \textbf{41.4}& \textbf{43.5}& \textbf{21.4}& \textbf{33.1}& \textbf{11.5}& \textbf{25.5}& \textbf{32.9}& \textbf{29.8}  \\ \hline
O & Oracle & \cmark & \xmark  & 38.7& 46.9& 56.7& 35.5& 49.4& 44.7& 35.9& 38.8& 43.1 \\ \hline
\end{tabular}}
\vskip-15.0pt
\end{table*}

\section{Experiments and Results}
\vskip-8.0pt
To validate the proposed method, we consider four domain shift scenarios where the source train model is adapted to the unlabelled target domain, typically used for comparison in UDA and SFDA literature. 
Specifically, we evaluate the proposed method with the existing UDA, SFDA and Test-time works under four domain shifts, 1) clear-weather to foggy-weather, 2) real to artistic, 3) synthetic to real, and 4) cross-camera adaptation. Note that, to show the effectiveness of our proposed approach, we evaluate both online and offline settings. Specifically, the offline setting follows the standard SFDA setting. The source-trained model is adapted towards the target domain using an unlabelled target train-set for multiple iterations and evaluated on the target test-set. Whereas in the online setting, the model is adapted towards the target domain in an online manner where the target test samples are seen only once and finally evaluated on the target test-set. This essentially simulates the real-world scenario where you see the target samples only once and adaptation needs to be continuous. 

\vspace{- 0.5 em}
\subsection{Implementation details}
\vspace{- 0.5 em}
For the Online adaptation setting, we adopt Faster-RCNN \cite{ren2015faster} with ResNet50 \cite{he2016deep} as the backbone pre-trained on ImageNet \cite{krizhevsky2012imagenet}.
In all of our experiments, the input images are resized with a shorter side to be 600 pixels while maintaining the aspect ratio.
We set the batch size to 1 for all experiments. For the student-teacher framework, the weight momentum update parameter $\alpha$ of the EMA for the teacher model is set equal to 0.99. The pseudo-labels generated by the teacher network with confidence greater than the threshold $T$=0.9 are selected for student training.
We utilize an SGD optimizer to train the student network with a learning rate of 0.001 and momentum of 0.9 for both online and offline training.
The Global Memory Bank contains $N_m$ memory items, which are set to 1024.
Further, the source model is trained using an SGD optimizer with a learning rate of 0.001 and momentum of 0.9 for 10 epochs. 
We report the mean Average Precision (mAP) with an IoU threshold of 0.5 for the teacher network on the distribution-shifted target domain test data during the evaluation.

\begin{table}
\caption{Quantitative results for Sim10K $\rightarrow$  Cityscapes and \\ KITTI $\rightarrow$  Cityscapes. }
\centering
\begin{center}
\label{tab:sim_kitti}
\resizebox{1.0\linewidth}{!}{\begin{tabular}{clcccc}
\hline
 Type  & Method  & Online & Ofline  & Sim10k $\rightarrow$ City & Kitti $\rightarrow$ City \\ \cline{5-6} 
                & &  &      & AP of Car      & AP of Car     \\ \hline
S  & Source-Only  & \cmark & \xmark   & 32.0          & 33.9  \\ 
 & DA Faster \cite{Chen2018DomainAF} (CVPR 2018)  & \cmark & \xmark   & 38.9          & 38.5   \\ 
 & MAF \cite{he2019multi} (ICCV 2019) & \cmark & \xmark  & 41.1 & 41.0\\
UDA & Robust DA \cite{khodabandeh2019robust} (ICCV 2019) & \cmark & \xmark  & 42.5& 42.9\\
 & Strong-Weak \cite{saito2019strong} (CVPR 2019) & \cmark & \xmark  & 40.1& 37.9\\
  & Harmonizing \cite{chen2020harmonizing} (CVPR 2020) & \cmark & \xmark   &  42.5 & -\\
  & Cycle DA \cite{zhao2020collaborative} (ECCV 2020) & \cmark & \xmark  & 41.5 &  41.7\\ 
  & MeGA CDA \cite{vs2021mega} (CVPR 2021) & \cmark & \xmark  & 44.8& 43.0\\
    & Unbiased DA \cite{deng2021unbiased} (CVPR 2021) & \cmark & \xmark  &43.1& -\\\hline
  & SFOD \cite{li2020free} (AAAI 2021) & \cmark & \xmark  & 42.3& 43.6\\
SFDA &  Mean-teacher \cite{tarvainen2017mean}  & \cmark & \xmark  & 42.3& 43.6\\
& MemCLR (Ours) & \cmark & \xmark  & \textbf{44.2}& \textbf{46.8}\\\hline
 O-SFDA & Tent \cite{wang2020tent} (ICLR 2021)  & \xmark & \cmark    & 32.8& 34.5 \\ 
    & MemCLR (Ours)   & \xmark & \cmark   & \textbf{37.2}& \textbf{38.5}  \\  \hline
\end{tabular}}
\end{center}
\vskip -25.0pt
\end{table}

\vspace{- 1.0 em}
\subsubsection{Clear-weather to foggy-weather adaptation} 
\vspace{- 0.5 em}
When the source-trained models are deployed in real-world applications such as autonomous navigation, they are likely to encounter data from multiple weather conditions such as fog, haze, etc.
In most cases, the deployed detector models would be trained for clear weather conditions.
We propose to formulate this as an online adaptation problem, as it is difficult to pre-determine what kind of weather conditions will occur.
Subsequently, we update the detector model in an online manner to adapt to any weather shifts the model might observe after deployment.
To evaluate the proposed method under such conditions, we experiment on Cityscapes \cite{cordts2016cityscapes} $\rightarrow$ FoggyCityscapes \cite{Sakaridis2018SemanticFS} dataset.
Here, we have a detection model trained on the Cityscapes dataset consisting of 2,975 normal weather images and 500 test images with 8 object categories: \textit{person, rider, car, truck, bus, train, motorcycle and bicycle}.
During inference, images from FoggyCityscapes are sequentially sent and the object detection model is adapted in an online manner to improve generalization on foggy/hazy weather. Table~\ref{tab:foggy} provides the comparison of the proposed FTTA method with the state-of-the-art UDA, SFDA, and O-SFDA methods for Cityscape$\rightarrow$FoggyCityscapes adaptation scenario.
From Table~\ref{tab:foggy}, we can infer that  UDA and SFDA methods operate in an offline manner, where as O-SFDA operates in an online manner. Firstly, in the online setting our proposed method outperforms existing UDA methods such as SWDA \cite{saito2019strong}, MTOR \cite{cai2019exploring} and InstanceDA \cite{wu2021instance} by a considerable margin. However, compared to MeGA-CDA \cite{vs2021mega} and Unbiased DA \cite{deng2021unbiased} our proposed method produces competitive performance with a drop of ~3-4 mAP. Note that these UDA methods have access to labelled source data, whereas under the SFDA setting, the proposed model only has access to source-trained model. Furthermore, the proposed method outperforms SFDA methods like SFOD \cite{li2020free} and HCL \cite{huang2021model} by 1.7 and 0.6 mAP, respectively. Secondly, when compared to the Test-time adaptation based methods such as Tent \cite{wang2021tent}, our best-performing model surpasses it by a huge margin of by 3.0 mAP. Therefore, for Cityscape$\rightarrow$FoggyCityscapes adaptation scenario, our proposed method produces state-of-the-art results in both online and offline SFDA settings. 

\vspace{- 1.0 em}

\begin{table}[t]
\caption{\small Quantitative results for PASCAL-VOC $\rightarrow$  Watercolor.\\\hspace{\textwidth}}
\label{tab:water}
\huge
\centering
\begin{center}
\resizebox{1.0\linewidth}{!}{\begin{tabular}{clccccccccc}
\hline
Type  & Method    & Online & Ofline      & bike          & bird        & car           & cat           & dog           & prsn          & mAP           \\ \hline

S  & Source only  & \cmark & \xmark  & 68.8          & 46.8          & 37.2          & 32.7          & 21.3          & 60.7          & 44.6          \\\hline
  & DA Faster \cite{Chen2018DomainAF} (CVPR 2018)  & \cmark & \xmark  & 75.2 & 40.6          & 48.0          & 31.5 & 20.6          & 60.0          & 46.0          \\ 
    & BDC Faster \cite{saito2019strong}  (CVPR 2019)  & \cmark & \xmark & 68.6          & 48.3 & 47.2          & 26.5          & 21.7          & 60.5          & 45.5          \\ 
 & BSR \cite{kim2019self}  (ICCV 2019) & \cmark & \xmark  & 82.8 & 43.2          & 49.8          & 29.6 & 27.6          & 58.4          & 48.6          \\ 
UDA & WST \cite{kim2019self}   (ICCV 2019) & \cmark & \xmark & 77.8 & 48.0          & 45.2          & 30.4 & 29.5          & 64.2          & 49.2          \\ 
  & SWDA \cite{saito2019strong}   (CVPR 2019) & \cmark & \xmark & 71.3 & 52.0          & 46.6          & 36.2 & 29.2          & 67.3          & 50.4          \\ 
  & HTCN \cite{chen2020harmonizing}  (CVPR 2020) & \cmark & \xmark  & 78.6 & 47.5          & 45.6          & 35.4 & 31.0          & 62.2          & 50.1          \\ 

 & $\text{I}^{3}$Net \cite{chen2021i3net} (CVPR 2021)  & \cmark & \xmark  & 81.1 & 49.3          & 46.2          & 35.0 & 31.9          & 65.7          & 51.5          \\ 
 & Unbiased DA \cite{deng2021unbiased} (CVPR 2021) & \cmark & \xmark   & 88.2 & 55.3          & 51.7          & 39.8 & 43.6          & 69.9          & 55.6          \\ \hline
  & SFOD \cite{li2020free}  (AAAI 2021) & \cmark & \xmark & \textbf{76.2}& 44.9& 49.3& \textbf{31.6}& 30.6& 55.2& 47.9  \\
SFDA  & Mean-teacher \cite{tarvainen2017mean} & \cmark & \xmark  & 73.6& 47.6& 46.6& 28.5& 29.4& 56.6& 47.1  \\ 
  & MemCLR (Ours) & \cmark & \xmark  & 70.7& \textbf{48.5}& \textbf{51.3}& 31.6& \textbf{34.0}& \textbf{61.3}& \textbf{49.6}  \\ \hline
O-SFDA  & Tent \cite{wang2020tent} (ICLR 2021) & \xmark & \cmark  & 62.3& \textbf{53.4}& 43.7& 29.5& \textbf{36.4}& 48.3& 45.4  \\ 
  & MemCLR (Ours) & \xmark & \cmark  & \textbf{66.1}& 46.2& \textbf{47.8}& 30.8& \textbf{30.0}& \textbf{55.3}& \textbf{46.1}  \\ \hline
\end{tabular}}
\end{center}
\vskip -25.0pt
\end{table}

\begin{figure*}[t]
\vskip+2.5mm
\begin{center}
\includegraphics[width=0.26\linewidth]{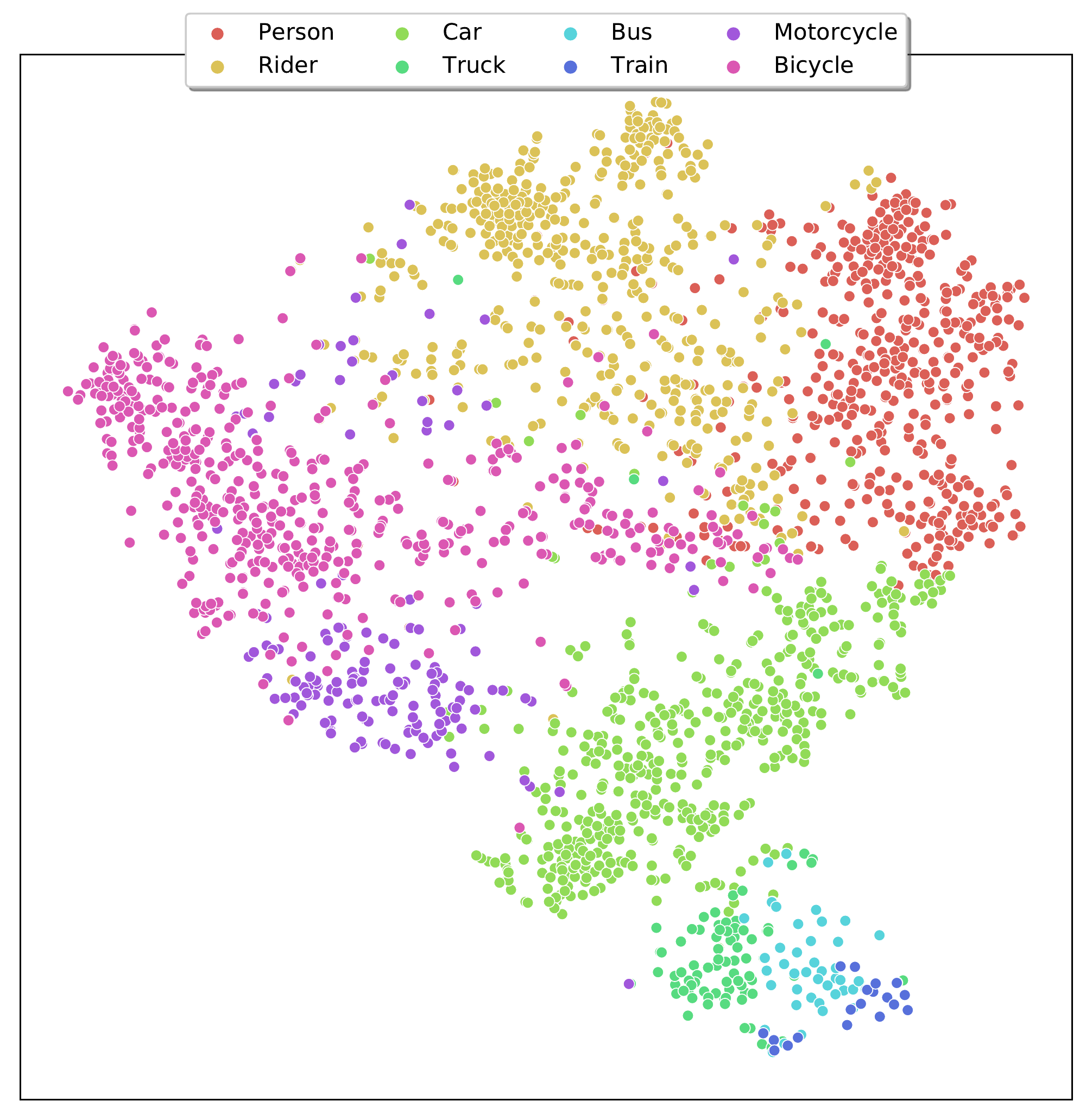}
\includegraphics[width=0.26\linewidth]{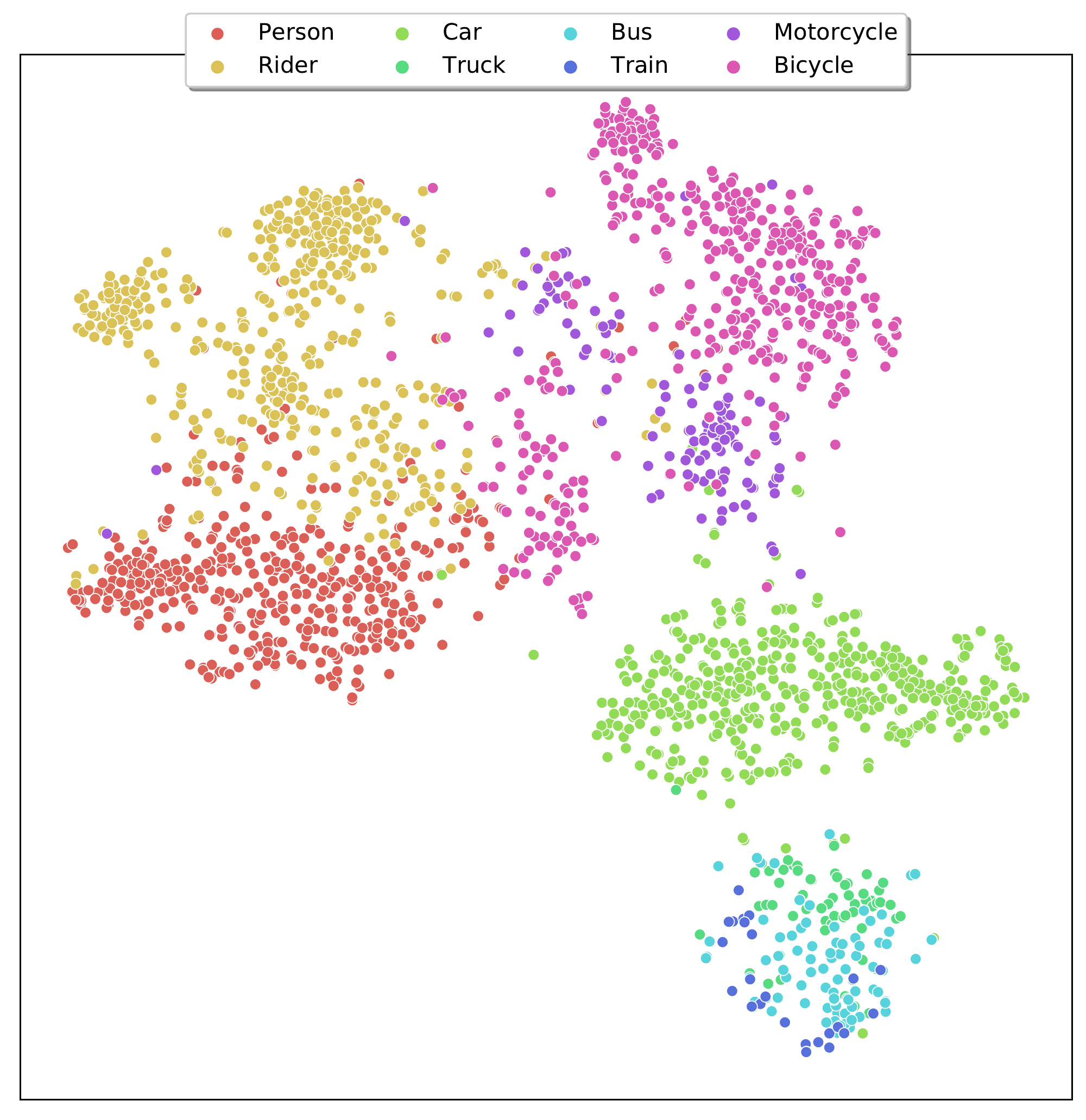}
\includegraphics[width=0.26\linewidth]{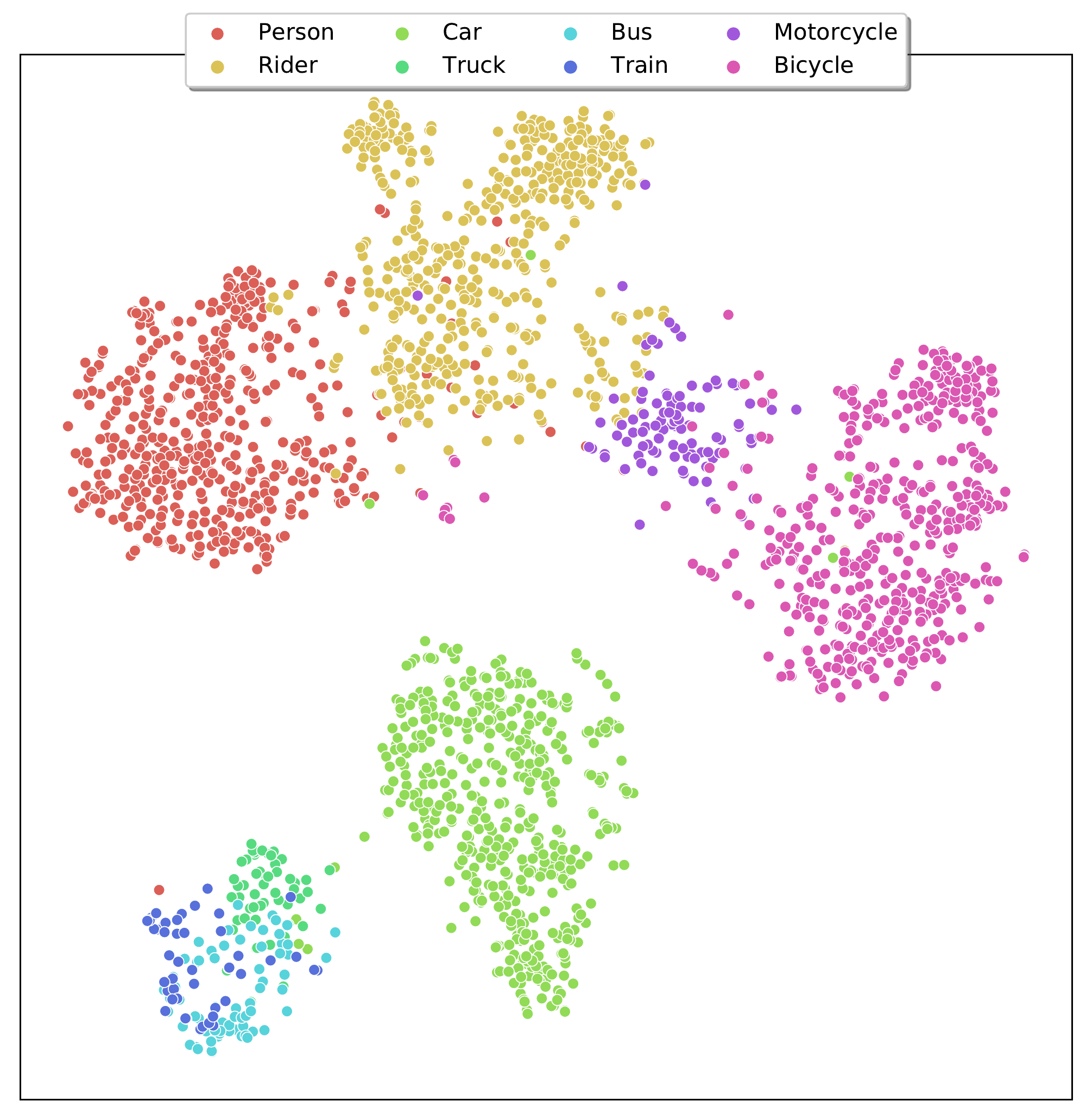}
\vskip -1pt
 \hskip -10pt (a) Source-only \hskip 55pt (b) Student-teacher  \hskip 75pt (c) Ours 
\end{center}
 \vskip -10.0pt\caption{ t-SNE \cite{van2008visualizing} visualization of RoI features for source-only, student-teacher and our methods for Cityscapes to FoggyCityScapes online setting. Different colors represent different classes. Compared to the source-only and student-teacher method, our proposed method has learned better classification boundaries  and compact feature representation for each category. }
\label{fig:tsne}
\vskip-15 pt
\end{figure*}

\begin{figure}[t]
\begin{center}
\includegraphics[width=0.32\linewidth]{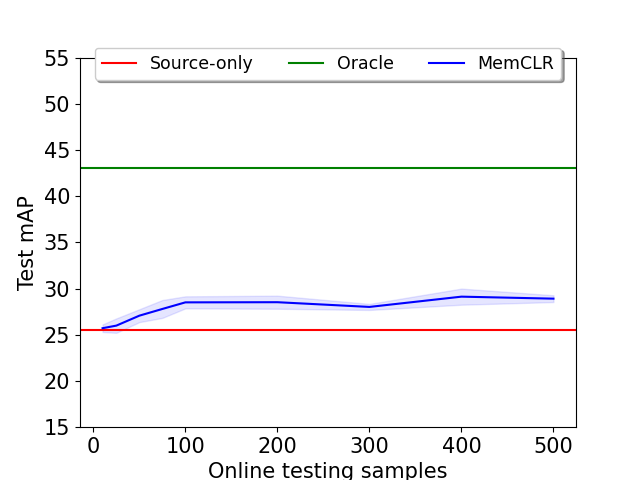}
\includegraphics[width=0.32\linewidth]{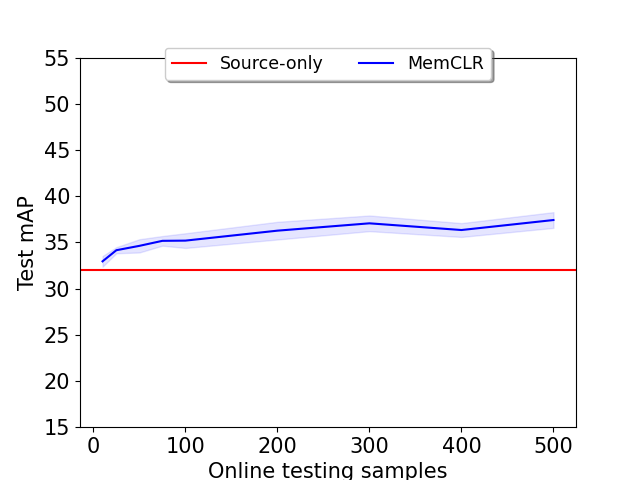}
\includegraphics[width=0.32\linewidth]{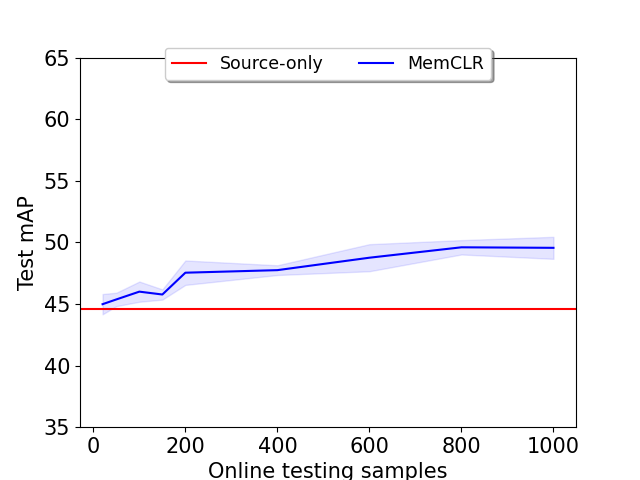}
\vskip -1pt
 \hskip -15pt (a) Foggy-Cityscapes \hskip 5pt  (b) Cityscapes  \hskip 15pt  (c) Watercolor 
\end{center}
 \vskip -10.0pt\caption{Quantitative comparison is performed to analyze the effect of the order of input sequence during online adaptation. We can observe from the variance that the input sequence order does not affect the model performance much. Note that, in online adaptation, the test samples are seen only once and adaptation happens in an unsupervised manner.  }
\label{fig:test_seq}
\vskip -20.0pt
\end{figure}

\subsubsection{Synthetic to real world adaptation}
\vspace{- 0.5 em}
Collecting and annotating detection data is computationally intensive, where on top of assigning a category, one needs to add bounding boxes to every object location in the image.
On the other hand, creating a synthetic dataset through simulation is much less computation-intensive and generates annotations for free.
Hence, training a detector model on a synthetically generated dataset makes sense and then deploying it in real-world conditions.
However, stylistic/appearance differences between real and synthetic data limit such deployment due to performance issues.
Here, we formulate it as an online adaptation problem to update a synthetic data trained model on the real-world test data.
Particularly, we consider a source model trained on Sim10k \cite{johnson2016driving} on 10,000 training images with 58,701 bounding boxes of \textit{car} category, rendered by the gaming engine \textit{Grand Theft Auto}.
For real-world test data we use the Cityscapes \cite{cordts2016cityscapes} validation set for online model adaptation. In Table~\ref{tab:sim_kitti}, we report Sim10K$\rightarrow$Cityscapes adaptation results on the existing UDA, SFDA, and O-SFDA methods. In an offline setting, compared to the existing UDA works such as DAFaster \cite{chen2018domain}, SWDA \cite{saito2019strong} and RobustDA \cite{khodabandeh2019robust}, the proposed method outperforms all of them by a considerable margin. Furthermore, when compared to SFOD \cite{li2020free} the proposed method is better by 0.7 mAP. In an online setting, compared to Tent \cite{wang2021tent}, our proposed method outperforms it by 4.0 mAP. Therefore, our proposed is able to perform well under synthetic to real-world domain shifts.
\vspace{- 0.5 em}
\subsubsection{Cross-camera adaptation} 
In most real-world applications, it is assumed that both training and test data would be collected using a camera with the same parameters.
However, the camera parameters are often different, which causes the collected images to have different appearances, such as radial distortions, tangential distortions, etc.
This can cause the model to perform poorly due to changes in the camera parameters.
Hence, to tackle any such camera distortions, we formulate the problem as an online adaptation problem and show that the proposed approach succeeds in generalizing to such cases.
Here, we have access to only the source model, trained on the KITTI \cite{geiger2013vision} dataset with 7,481 training images with bounding boxes for the  \textit{car} category.
To emulate cross-camera scenario, we consider online adaptation on the Citsycapes \cite{cordts2016cityscapes} validation set containing 500 images. We report the results of the cross-camera adaptation experiment in Table~\ref{tab:sim_kitti}. Similar to Sim$\rightarrow$Cityscapes adaptation even for Kitti$\rightarrow$Cityscapes adaptation, we show similar performance improvements compared to UDA, SFDA and O-SFDA methods.
Specifically, in the O-SFDA setting, the proposed method outperforms Tent \cite{wang2021tent} by 5.6 mAP. Thus, our proposed method is able to model the cross-camera domain shifts effectively.

\vspace{- 0.5 em}

\subsubsection{Real to artistic adaptation} 
Here, we evaluate the proposed method for the case where there is a \textit{concept shift} in during inference.
By \textit{concept shift}, we refer to the case where there is a complete change in the object, e.g., going from real-world to artistic images.
Unlike previous scenarios where the objects go through stylistic/appearance changes, the entire \textit{concept} of an object is different, e.g., a real-world car vs a cartoon car \cite{inoue2018cross}.
We show that even in this challenging scenario, the proposed approach is able to improve model generalization through online updates.
We consider a model trained on the Pascal-VOC data \cite{everingham2010pascal} which adapts to test set of  Watercolor \cite{inoue2018cross}. 
Specifically, the Watercolor consists of 1K training and 1K testing images with six categories.
 We compare PASCAL-VOC$\rightarrow$Watercolor results with the existing methods in Table~\ref{tab:water}. From Table~\ref{tab:water}, we can infer that the proposed method outperforms most of the existing UDA methods and SFDA methods in offline settings. Further, in the online setting, when compared to TENT\cite{wang2020tent} the proposed method is able to outperform by a significant margin. This demonstrates the capability of the proposed method to generalize even for both online and offline settings. 

\begin{table}[t]
\caption{Ablation analysis on Cityscapes$\rightarrow$FoggyCityscapes.}
\label{tab:foggy_ab}
\centering
\huge
\resizebox{0.8\linewidth}{!}{\begin{tabular}{lcccccccccc}
\hline
Method     & Mem items   & prsn          & rider         & car           & truck         & bus           & train         & mcycle        & bcycle       & mAP           \\ \hline
\hline
Source-only & \xmark & 29.3          & 34.1          & 35.8          & 15.4          & 26.0          & 9.09          & 22.4          & 29.7          & 25.2          \\ \hline
Student-Teacher & \xmark & 33.1  & 42.2 & 44.7 & 24.0& 33.6 & 17.8& 26.8& 38.1 & 32.5  \\ 
SupCon   & \xmark & 33.0& 43.1& 49.8 & 26.5& 31.1& 23.3& 27.7& 37.2& 33.8 \\ 
MemCLR (Ours)   & 256  & 37.2& 41.7& 51.3& \textbf{27.5}& 38.5& 28.5& 29.6& 39.3& 36.7  \\ 
MemCLR (Ours)  & 512  & 37.4& \textbf{45.2}& 51.9& 24.4& 39.6& 25.2& \textbf{31.5}& 41.6& 37.1  \\ 
MemCLR (Ours)   & 1024  & \textbf{37.7}& 42.8& \textbf{52.4}& 24.5& \textbf{40.6}& \textbf{31.7}& 29.4&\textbf{ 42.2}& \textbf{37.7}  \\ \hline
\end{tabular}}
\vskip -15.0pt
\end{table}

\subsection{Ablation analysis}
\vskip -2pt
\noindent\textbf{Quantitative analysis.}
The Cityscapes$\rightarrow$FoggyCityscapes ablation experiment results are reported in  Table~\ref{tab:foggy_ab} for the offline-SFDA setting.
We first consider a student-teacher offline update baseline which, compared to the source-only baseline, provides significant improvements.
To have a fair comparison, we also consider utilizing supervised contrastive loss \cite{khosla2020supervised} for offline updates.
In particular, we utilize predictions provided by student-teacher training as label information needed for applying the supervised contrastive loss over object proposals.
Denoted as SupCon in Table~\ref{tab:foggy_ab}, the addition of supervised contrastive learning further improves the performance by 1.3 mAP.
However, the proposed memory-based contrastive learning outperforms the supervised contrastive learning by 1.4 mAP, indicating the utility of the proposed method to learn better target representations.
Finally, we analyze the performance of the proposed method by varying global memory bank capacity from 256 to 1024 memory items.
As shown in Table~\ref{tab:foggy_ab}, memory-based contrastive loss with 1024 memory items performs the best when compared to 256 and 512 memory items. Further, note that our model takes around 1 second to perform online adaptation for one sample. 

\noindent\textbf{Qualitative analysis.}  Fig.~\ref{fig:tsne} shows t-SNE visualization for source-only, student-teacher training and the proposed method for the  Cityscapes$\rightarrow$FoggyCityscapes online-SFDA setting.
The t-SNE \cite{van2008visualizing} visualizations are created from the RoI features extracted from the predictions for 500 test images.
Due to the distribution shift, the features are dispersed for the source-only baseline and classification boundaries are weak.
With the help of student-teacher training, the model learns better classification boundaries, resulting in better quantitative performance.
However, the features in the student-teacher training have a large variance and do not have compact features.
Whereas the proposed method has even better classification boundaries and learns compact features for each category, resulting in a more robust model. Further qualitative comparison is performed to analyze the effect of the order of input sequence during online adaptation is shown in Fig. \ref{fig:test_seq}. Multiple experiments with changing the order of input sequence are conducted and corresponding performance mean and variance is plotted in Fig. \ref{fig:test_seq}. We can observe from the variance that the order of input sequence does not much affect the model's performance. Further, we can observe the model performance increase as it encounters more test samples during online adaptation, showing the MemXformer effectiveness in exploiting online target distribution. Note that, in online adaptation, the test samples are seen only once and adaptation happens in an unsupervised manner. 

\section{Conclusion}

In this work, we introduced a practical domain adaptation setting for the object detection task, which is feasible for real-world settings.
Particularly, we proposed a novel unified adaptation framework which makes the detector models robust against online target distribution shifts. Further, We introduce the MemXformer module, which stores prototypical patterns of the target distribution and provides contrastive pairs to boost the contrastive learning on the target domain. 
We conducted extensive experiments on multiple detection benchmark datasets and compared existing unsupervised domain adaptation, source-free domain adaptation and test-time adaptation methods to show the effectiveness of the proposed approach for both online and offline adaptation of object detection models. We also analyzed multiple aspects of the proposed method in ablation experiments and identified increasing the online adaptation speed further is a potential directions for future research.

{\small
\bibliographystyle{ieee_fullname}
\bibliography{egbib}
}

\end{document}